\documentclass[10pt,twocolumn,letterpaper]{article}

\usepackage{iccv}
\usepackage{times}
\usepackage{epsfig}
\usepackage{graphicx}
\usepackage{amsmath}
\usepackage{amssymb}
\usepackage{algorithm}
\usepackage{algorithmic}
\usepackage{authblk}

\usepackage[pagebackref=true,breaklinks=true,letterpaper=true,colorlinks,bookmarks=false]{hyperref}

\iccvfinalcopy 

\def\iccvPaperID{5} 

\ificcvfinal\fi

\begin{document}

\title{ Disjoint Pose and Shape for 3D Face
Reconstruction}


\author{Raja Kumar$^*$ \qquad Jiahao Luo$^*$ \qquad Alex Pang \qquad James Davis \\University of California Santa Cruz \\ \texttt{\{rkumar38, jluo53, pang, davisje\}@ucsc.edu}}

\maketitle
\def\thefootnote{*}\footnotetext{authors contributed equally}

\ificcvfinal\thispagestyle{empty}\fi

\begin{abstract}
Existing methods for 3D face reconstruction from a few casually captured images employ deep learning based models along with a 3D Morphable Model(3DMM) as face geometry prior. Structure From Motion(SFM), followed by Multi-View Stereo (MVS), on the other hand, uses dozens of high-resolution images to reconstruct accurate 3D faces. However, it produces noisy and stretched-out results with only two views available. In this paper, taking inspiration from both these methods, we propose an end-to-end pipeline that disjointly solves for pose and shape to make the optimization stable and accurate. We use a face shape prior to estimate face pose and use stereo matching followed by a 3DMM to solve for the shape. The proposed method achieves end-to-end topological consistency, enables iterative face pose refinement procedure, and show remarkable improvement on both quantitative and qualitative results over existing state-of-the-art methods.


\end{abstract}

\section{Introduction}
\label{intro}
3D face models find a wide range of applications in scenarios such as 3D avatars \cite{zollhofer2011automatic}, biometric identification  \cite{blanz2003face}, photo editing \cite{yang2011expression} and film production \cite{borshukov2005universal}. Traditionally, specialized setups and hardware \cite{debevec2000acquiring, borshukov2005universal} have been used to generate high-fidelity 3D faces. However, reconstructing accurate 3D faces from a few casually captured uncalibrated images remains a challenging problem. \
\begin{figure}[h!]
  \centering
    \includegraphics[width=\linewidth]{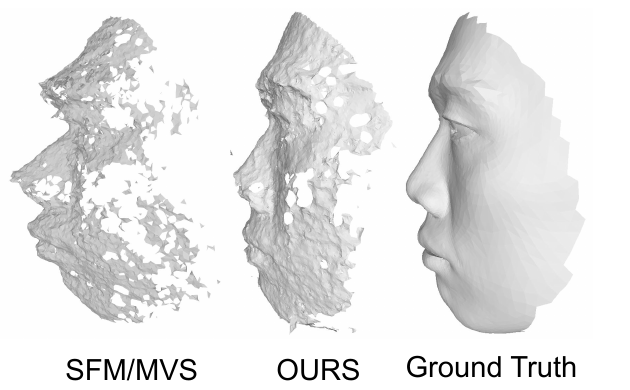}
    \caption{Comparison of raw stereo output from SFM/MVS \cite{schonberger2016pixelwise} \cite{schoenberger2016sfm} (left), our method (middle), and ground truth (right). Clearly, SFM/MVS produces a stretched-out face with more noise due to under-constrained optimization with only 2-views. The proposed face shape prior solves this problem by providing a strong prior.}
    \label{fig:figure6}
\end{figure}

Recent learning based methods use a Deep Neural Network (DNN)  model to reconstruct 3D faces. Most often, a 3D Morphable Model (3DMM) is employed to represent the face shape using a vector space representation that is learned from a linear combination of principle components of a collection of face scans. These methods aim to recover the 3DMM parameters from given facial images, often by analysis-by-synthesis optimization or by employing multi-view constraints. However, it requires a complicated, nonlinear optimization that has difficulty converging in practice. 
\begin{figure*}
\centerline{\includegraphics[width=1\linewidth]{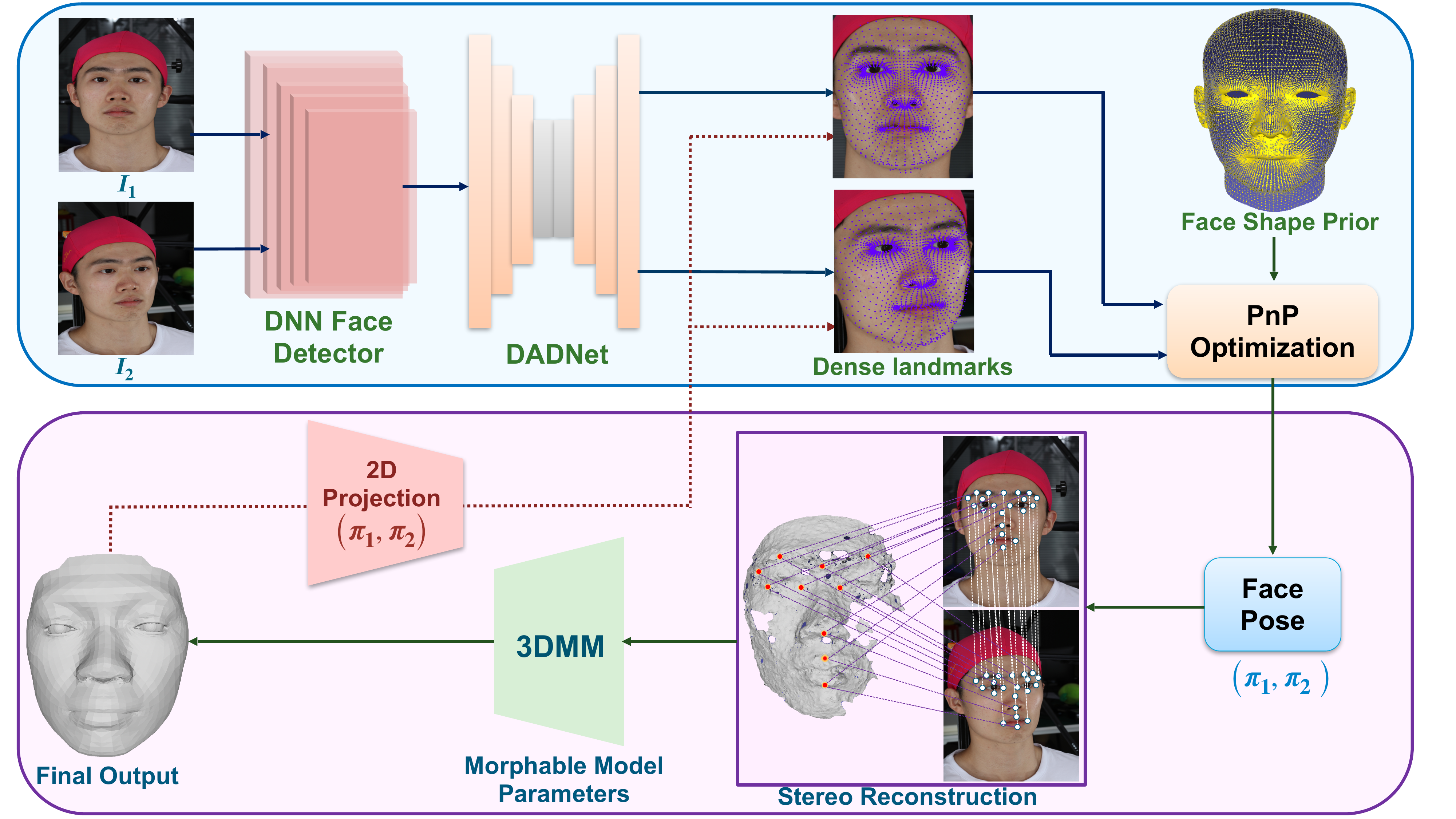}}
\caption{An overview of the proposed method. In the proposed method, we solve for pose and shape disjointly for an accurate reconstruction. The blue box shows the proposed face pose estimation using face shape prior method, and pink box shows the 3D face reconstruction pipeline. $\mathbf{\pi_1} \, and \, \mathbf{\pi_2}$ represent the projection matrices.}
\label{figure:fig1}
\end{figure*}

In contrast, the highest quality 3D faces are generated using multi-view stereo methods with dozens of calibrated high-resolution images captured in a laboratory setting.  Unfortunately pre-calibrated camera pose is not readily available in a casual capture setting. Recent learning-based methods consisting of Structure From Motion (SFM) followed by Multi-View Stereo (MVS) have been proposed for 3D reconstruction from uncalibrated sets of images. However, solving for camera poses using this method when only a few views are available are under-constrained and results in inaccurate estimate as shown in Figure \ref{fig:figure6} above.


The key observation in this paper is that when only two views are available jointly optimizing for pose parameters, shape parameters, and correspondence between image and model is theoretically optimal, but results in sub-optimal results in practice. Instead we propose a disjoint solution in which pose and shape are solved separately. This allows appropriate regularization priors to be used in each stage, allowing a more stable overall solution.


Our end-to-end pipeline consists of three stages, face pose estimation using a face shape prior, 3D reconstruction using stereo matching, and iterative camera pose refinement. In the first stage, given two views, we detect topologically consistent dense 2D landmarks for both views and use a strongly constrained 3D face shape prior in the same topology as the dense 2D landmarks. Given 2D landmarks and the 3D face shape prior, we solve for camera poses using Quadratically Constrained Quadratic Program (QCQP) based optimization proposed in Terzakis \etal \cite{terzakis2020consistently}.  In the second stage of our pipeline, we perform stereo matching to generate a 3D point cloud. This step uses no face shape prior in order to allow a full range of shape variation. This is followed by a fit to a modern 3DMM model called FLAME to fill in the missing region. The 3DMM acts as a face prior on the raw 3D point cloud, however it allows a great deal of variation in shape and is thus much less strongly constrained than the face prior used to find pose.  

Since our output accuracy is determined in part by the estimate of pose derived from 2D landmarks, we perform an iterative refinement on these. The 2D landmarks are refined using the current estimates of pose and shape. This refinement converges in only a few iterations. Our analysis on FaceScape and Stirling datasets show that the proposed pipeline outperforms the state-of-the-art multi-view methods in quantitative as well as qualitative comparison.




The contribution of this paper is an end-to-end pipeline for 3D face reconstruction by solving pose and shape disjointly using two uncalibrated images, achieving state-of-the-art accuracy. 

\section{Related Work}
\label{realted_work}
\subsection{3D Morphable Face Model}
\label{3DMM}
3DMM is a statistical model which transforms the shape and texture into a vector space representation that is derived from hundreds of 3D face scans. It was first proposed by Blanz \etal ~\cite{blanz1999morphable}, and numerous variations such as~\cite{paysan20093d, li2017learning} were introduced to include identity, expression, and pose factors. This allowed for separate control of the model, and using these attributes, it can be transformed into a set of blendshapes~\cite{li2010example} which can then be rigged to create unique animations for each individual. More recent deep learning based methods ~\cite{bagautdinov2018modeling, tran2019towards, tran2019learning, tran2018nonlinear} have been proposed to enhance the representation power of 3DMM. We recommend referring to the recent survey~\cite{egger20203d} for a comprehensive review. In this paper, we use the
FLAME model~\cite{li2017learning} for its simplicity and wide applicability.

\begin{figure*}

\centerline{\includegraphics[width=1\linewidth]{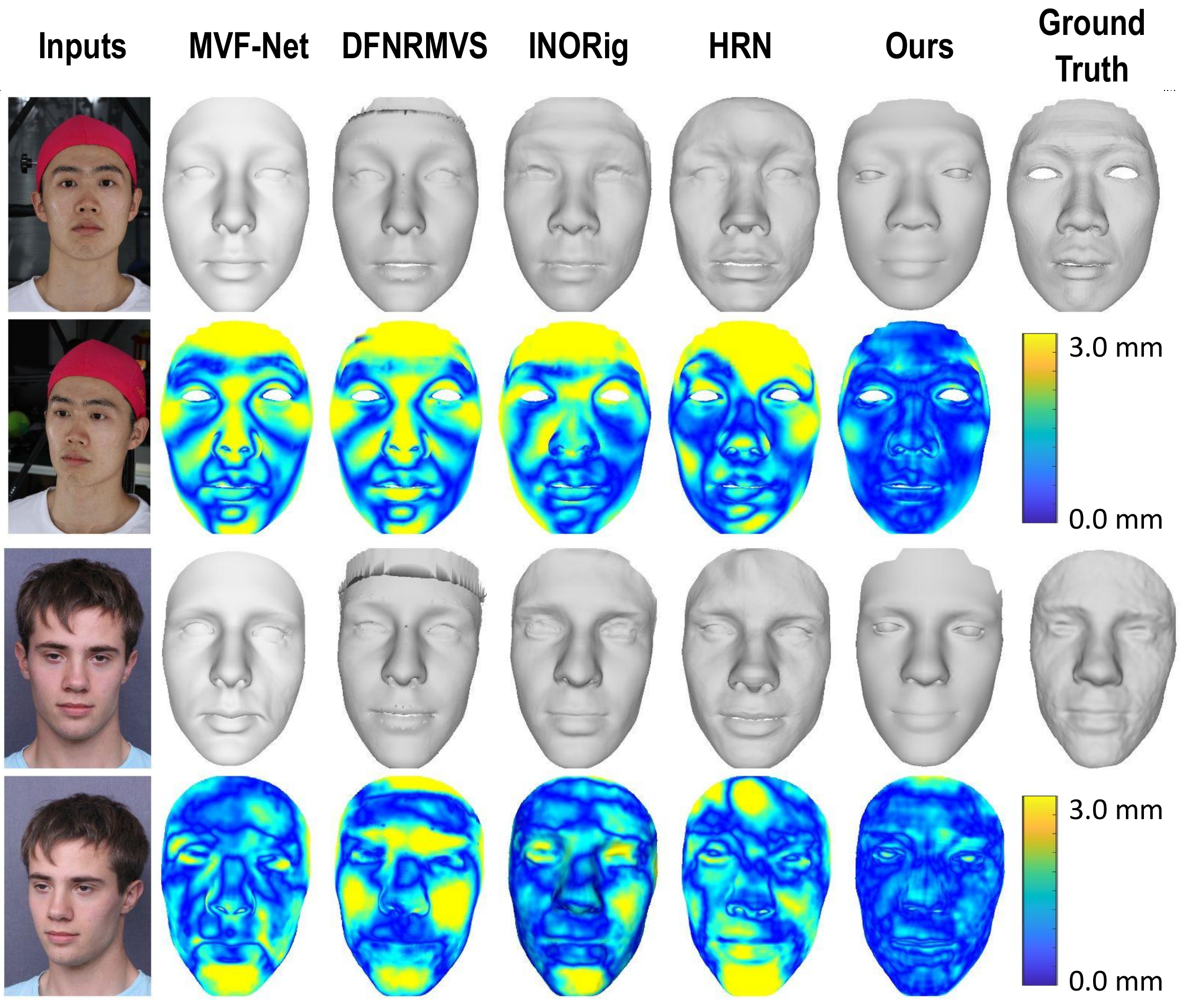}}
\caption{A qualitative comparison of output rendering and error map to the SOTA methods. The first and second rows show the output and error map on a sample input from the FaceScape dataset, while the third and fourth show the output and error map on a sample of the Stirling dataset. The error maps visualize a range from blue (0 mm) to yellow (3 mm). It is evident that the proposed method produces a smooth reconstruction with lower error.}

\label{figure:fig4}
\end{figure*}

\subsection{Multi-View Stereo}
\label{MVS}
Traditional MVS methods~\cite{furukawa2009accurate, kutulakos2000theory, lhuillier2005quasi} estimate the 3D shape of the object from a set of calibrated multi-view images. These methods perform feature matching, followed by triangulation, and then the application of a depth map fusion algorithm to obtain 3D meshes~\cite{furukawa2009accurate, campbell2008using, schonberger2016pixelwise, tola2012efficient}. Parts of this pipeline that solve for specific missing information or improve the speed and accuracy of reconstructions have also been proposed. Depth fusion methods have been used to produce a valid 3D watertight surface~\cite{curless1996volumetric, merrell2007real, kazhdan2006poisson}. Multiple stereo-based methods~\cite{valgaerts2012lightweight, beeler2010high} have been proposed specifically for faces. Valgaerts \etal ~\cite{valgaerts2012lightweight} proposed a lightweight passive facial performance capture approach that uses image-based scene flow computation, lighting estimation, and shading-based refinement algorithms to reconstruct the face. Beeler \etal ~\cite{beeler2010high} used a combination of smoothness, ordering, and uniqueness constraints to recover facial shape robustly.  However, these methods generally require pre-calibrated cameras or many images. Our work focuses on face reconstruction from two uncalibrated images.

\subsection{Deep Face Reconstruction}
\label{DFR}

 More recent learning-based methods~\cite{hartmann2017learned, chen2019point, yi2020pyramid} train DNN with different objective functions to improve 3D face reconstruction. Other methods~\cite{dou2018multi, ramon2019multi, bai2020deep} proposes to recover facial geometry from uncalibrated images or videos. Duo \etal ~\cite{dou2018multi} uses DNN to disentangle facial and identity features and uses Recurrent Neural Network (RNN) to perform subspace representation of 3D shapes. Ramon \etal ~\cite{ramon2019multi} uses a Siamese network to extract features from multi-views. Wu \etal ~\cite{wu2019mvf}  incorporates multi-view geometric constraints into the network by establishing dense correspondences between different views leveraging a novel self-supervised view alignment loss. Though these methods achieve good results, they require multiple views, and performance is sub-optimal when only two views are available. 
 
 3D face reconstruction from a single view has also been studied and is challenging due to its ill-posed nature. Generally, a deep neural network is trained to regress 3DMM model parameters~\cite{tuan2017regressing, zhu2016face} reconstruct 3D geometries~\cite{guo2020towards, li2017learning, sanyal2019learning} or render image using analysis-by-synthesis~\cite{deng2019accurate, gecer2019ganfit, feng2021learning}. Prior work~\cite{luo2022much, wu2019mvf} has shown that multi-view face methods generally perform better than single-view. In this paper, we compare directly with several state of the art multi-view reconstruction methods. 

\section{Proposed Method}
\label{prop_meth}

The proposed method consists of three stages, as shown in Figure \ref{figure:fig1}. In the first stage, we estimate the face poses using topologically consistent dense 2D landmarks and a 3D face shape prior. In the second stage, an accurate 3D face is reconstructed using two-view stereo matching, which utilizes the face pose obtained in the first stage. This is followed by 3DMM model fitting to fill in the missing region. Finally, we perform Face Pose Refinement (FPR) by iteratively projecting the 3D face to image space using the face pose estimated in the previous iteration. We explain each stage in detail in the following sections.

\subsection{Pose Estimation Using Face Shape Prior}
\label{sec3_1}
In the first stage of our pipeline, we use topologically consistent dense 2D landmarks and a 3D
face shape prior for face pose estimation. In this section, first we provide details about dense 2D landmark detection and face shape prior computation, followed by an explaination of the face pose estimation method.

Facial landmark detection is a well-explored field of research. However, most of the existing methods focus on a set of frequently used 68 landmarks. Martyniuk \etal \cite{dad3dheads} and Wood \etal \cite{wood20223d} proposes dense landmark detection methods with 10x more landmark points. We first use a face detector  \cite{bazarevsky2019blazeface} to find the facial region since landmark detection often fails to find accurate landmarks without it.  DADNet \cite{dad3dheads} is used to find dense landmarks. 

We use a face shape prior as a regularizer to solve for face poses. We use the mean face with no variability to provide a very strong prior. Given a set of $m$ training set faces $\mathcal{F} = \{\mathbf{F_1, F_2, ... , F_m}\}$, we compute the face shape prior $\mathbf{F_p}$ as the mean of all the face vertices in correspondence
\begin{equation}
    \label{eq2}
    \mathbf{F_p}^k = \frac{1}{m} \sum_{i=1}^m \mathbf{F_i}^k \hspace{1cm} \forall k \in \{1,2,...,n\}
\end{equation}

where, $\mathbf{F_p}^k$ and $\mathbf{F_i}^k$ denote the $k^{th}$ vertex of face prior $\mathbf{F_p}$ and the $i^{th}$ face $\mathbf{F_i}$ respectively.

\begin{algorithm}
\caption{Iterative FPR and 3D Reconstruction}
\label{algo1}
\textbf{Input:} Two view images ($I_1, I_2$), face shape prior ($F_p$) computed as explained in Section \ref{sec3_1} \\
\textbf{Output:} Reconstructed 3D face and Face Poses
\begin{enumerate}
    \item \textbf{Init:} $L_1 = LD(I_1)$, $L_2 = LD(I_2)$,  where, $L_1, L_2 \in \mathbb{R}^{Nx2}$ and LD is Landmark Detector
    \item \textbf{for } $iter=0,1,\dots,$till convergence
    \begin{enumerate}
        \item ($R_j, t_j$) $= argmin(R,t) \sum_{i=1}^n || p_i - K(Rv_i+t)||^2$  \hspace{0.2cm} $\forall p_i \in L_j, v_i \in F_p  \, and \, j \in \{1,2\}  $
        \item $F_{sparse} = Stereo(I_j, R_j, t_j)$ \hspace{0.5cm} $\forall j \in \{1,2\}$
        \item $F_{flame} = 3DMM(F_{sparse})$
        \item $L_j = K(R_jF_{flame}+t_j)$ \hspace{0.5cm} $\forall j \in \{1,2\}$  \hspace{1.0cm}// 2D projection of 3D face
        \item Next iteration with update landmarks ($L_j$)
    \end{enumerate}
\end{enumerate}
\end{algorithm}

Face pose estimation requires solving for the rotation $\mathbf{R} \in$ SO(3) and translation $\mathbf{t} \in \mathbb{R}^3 $ for an image $I$. However, solving for face poses using only two views when the face shape is not yet known is a highly under-constraint problem and has shallow minima leading to many solutions. Hence, to constrain our hypothesis space, we use the face shape prior.  Formally, Given an image $I$, we find the topologically consistent landmark points $\mathbf{L} = \{\mathbf{p_1, p_2, ... , p_n\}} \in \mathbb{R}^{nx2}$ and face shape prior $\mathbf{F_p} = \{\mathbf{v_1, v_2, ... , v_n}\} \in \mathbb{R}^{nx3}$ and we seek to find the rotation $\mathbf{R}$ and $\mathbf{t}$ minimizing the cumulative squared projection error

\begin{equation}
\label{eq1}
    S(\textbf{R,t}) = \sum_{i=1}^n || \mathbf{p_i} - \mathbf{K \cdot (R \cdot } \mathbf{v_i}+\mathbf{t})||^2 = \sum_{i=1}^n || \mathbf{p_i} - \mathbf{\pi} \cdot v_i||^2
\end{equation}

Where, $\mathbf{R, t}$ are the rotation and translation parameters to be estimated, $\mathbf{K}$ is the scaling factor using weak perspective projection, $\pi$ is the projection matrix and $\mathbf{p_i}$ and $\mathbf{v_i}$ represent the $i^{th}$ points in the image and 3D face prior space respectively. This is a well-known Perspective-n-Point (PnP) optimization problem, and we use a Quadratically Constrained Quadratic Program (QCQP) based implementation proposed in Terzakis \etal \cite{terzakis2020consistently}, to solve this optimization problem. Figure \ref{figure:fig1} (blue box) shows our face pose estimation pipeline. 

\begin{figure*}
    \centerline{\includegraphics[width=1\linewidth]{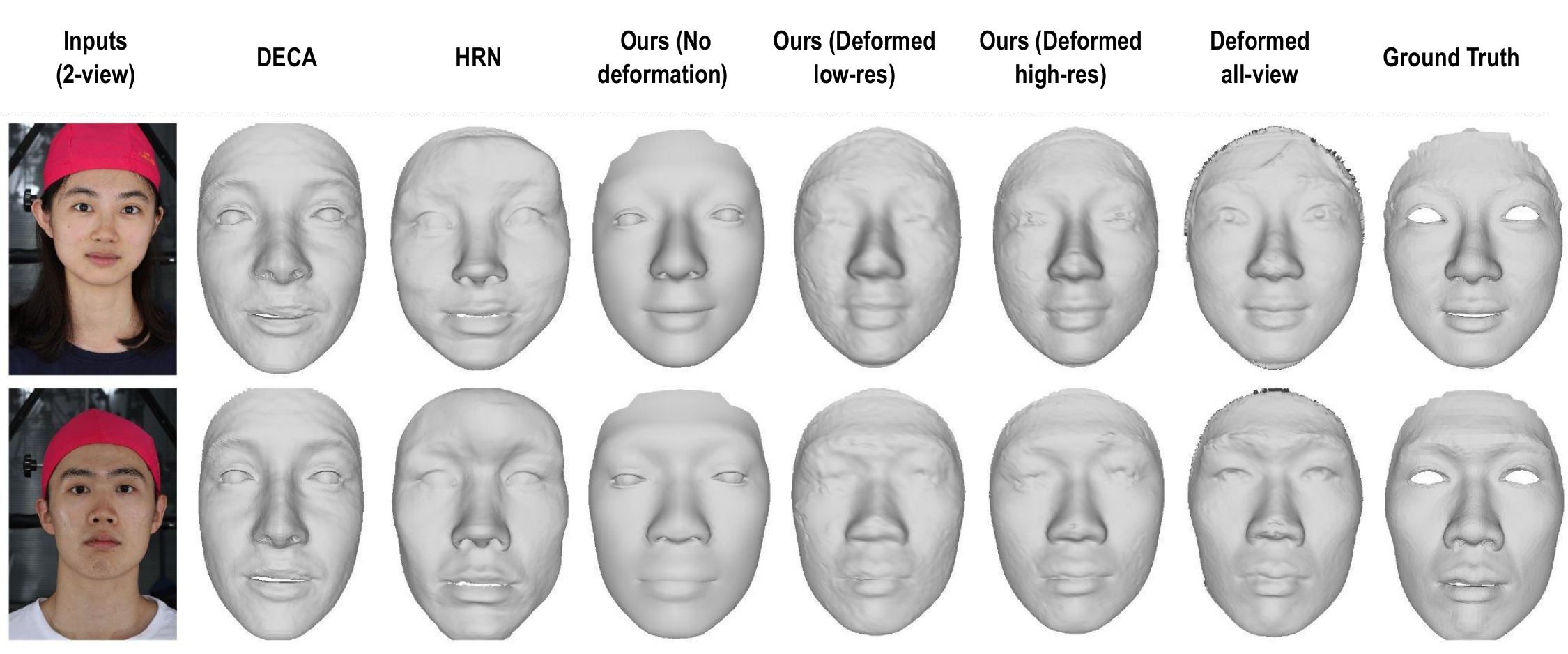}}
    \caption{Our dense intermediate stereo output can be used to generate shape refinement beyond 3DMM. A qualitative comparison of deformed rendering to SOTA methods on FaceScape \cite{yang2020facescape} dataset. (from left to right) DECA \cite{feng2021learning}, HRN \cite{lei2023hierarchical}, ours without deformation, our deformed outputs, deformed outputs with high resolution inputs, deformed outputs with 50-view inputs, and the ground truth}
    \label{figure:facescape_deform}
\end{figure*}

\subsection{3D Face Reconstruction}
\label{sec_3_2}
Given two images and the associated face poses estimated using the method described in the previous section, we use the classical stereo method for 3D reconstruction. We do not employ any face shape prior during stereo matching. This allows maximum shape variability, however it produces noisy results and leaves gaps in areas not visible to both cameras. 

To obtain a complete 3D facial model, a statistical blendshape 3D face model called FLAME is utilized as the geometry prior. This model consists of the head, neck, eyeball, and shoulder region and was developed from more than 33,000 scans. Despite its low dimensionality, the FLAME model is considered more expressive than other models, such as the FaceWarehouse \cite{cao2013facewarehouse} and Basel Face Model \cite{li2017learning}. The model's parameters include both shape and expression, and this prior allows significantly more variation while solving for shape than the static prior used when solving for pose.  The authors' original implementation of FLAME is used to determine the best-fitting parameters for the available data. 

\begin{figure}
    \centerline{\includegraphics[width=1\linewidth]{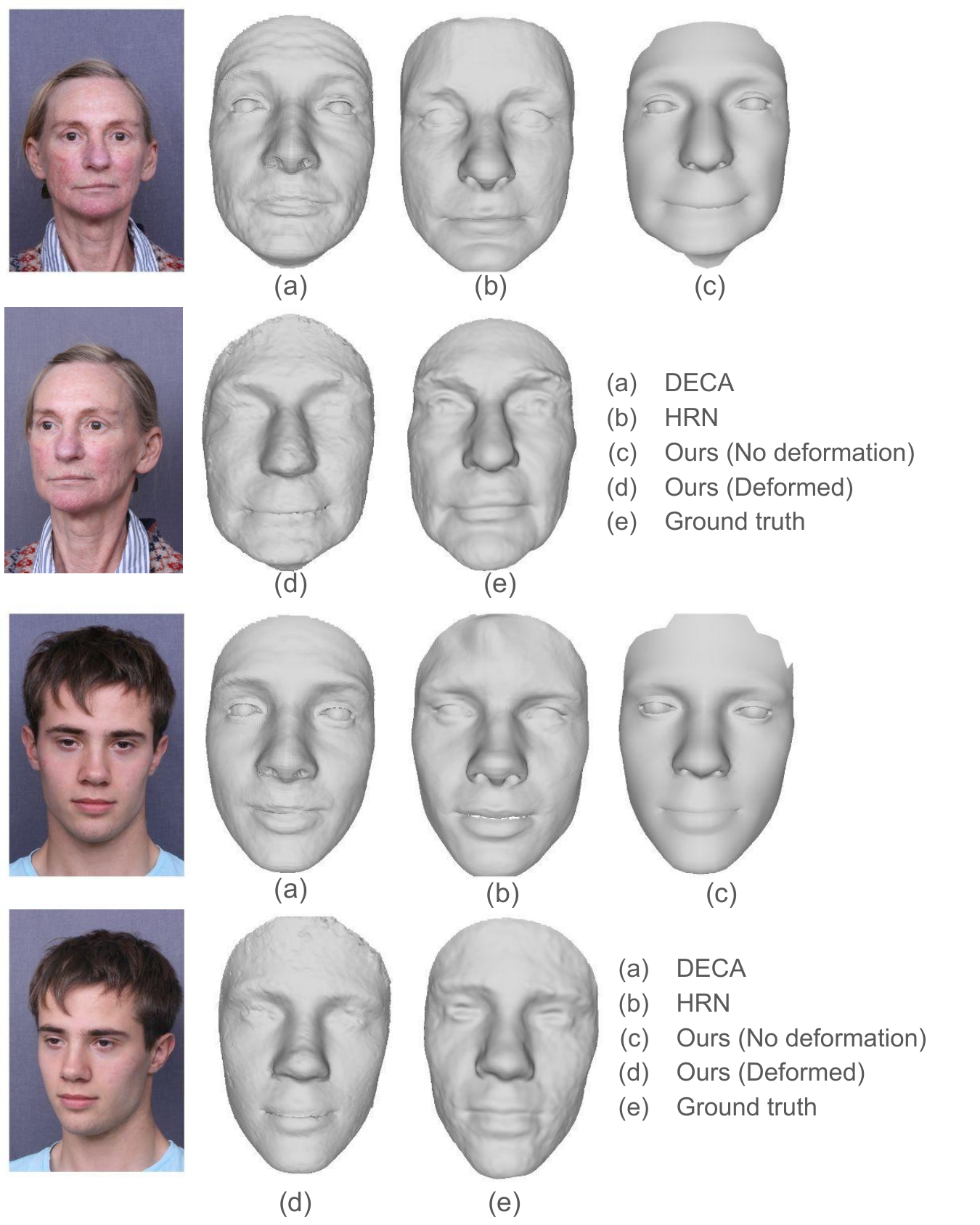}}
    \caption{ A qualitative comparison of
deformed rendering to SOTA methods on Stirling \cite{Stirling2018} dataset. Our method can recover mid-frequency details which matches with groudtruth.}
    \label{figure:stirling_deform}
\end{figure}

\begin{figure*}
\begin{minipage}[h]{0.47\linewidth}
\begin{center}
\includegraphics[width=0.95\linewidth]{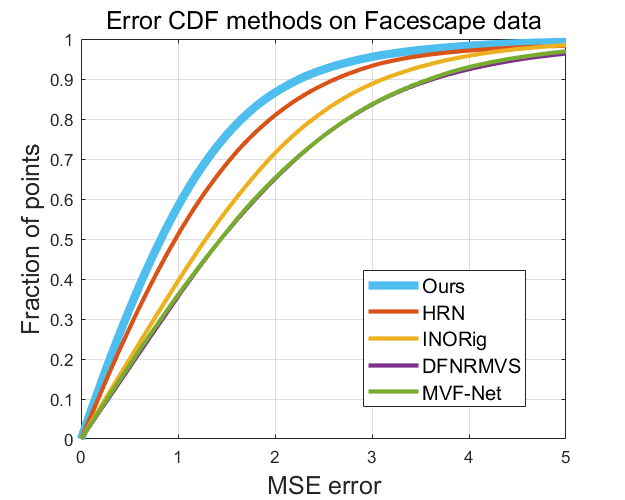} 
\end{center}
\end{minipage}
\hfill
\begin{minipage}[h]{0.47\linewidth}
\begin{center}
\includegraphics[width=0.98\linewidth]{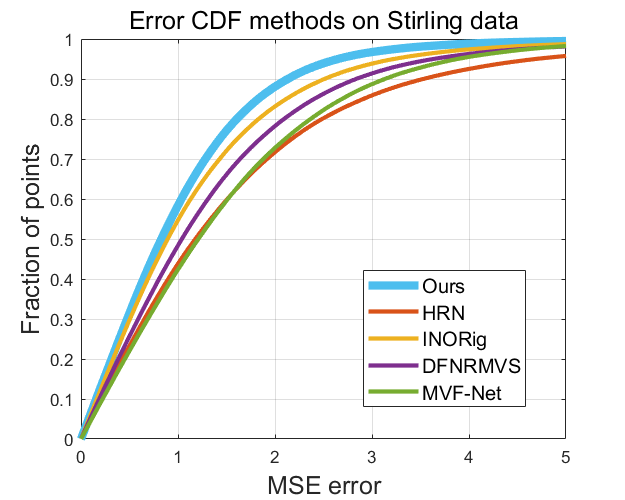} 
\end{center}
\end{minipage}
\vspace{0.3 cm}
\caption{Comparison of performance plotted as a CDF with error on the x-axis and the fraction of reconstructed points with the error below this bound on the y-axis.Our proposed method has accuracy substantially better than the existing methods. A horizontal
line through the plot at 0.9 is identical to the M90
error metric reported in results.}
\label{figure:fig3}
\end{figure*}

\subsection{Iterative Face Pose Refinement}

In the proposed pipeline, we find the 2D landmarks for two views independently using DADNet for initialization. However, the two views constraints introduced using stereo can help improve these landmarks, and consequently, the face poses. To this end, we perform an iterative Face Pose Refinement (FPR). Once we obtain the 3D face and face pose in the first iteration, we perform 2D projection and pick the corresponding landmarks. This is possible because the landmarks of our 2D detector are topologically consistent with the vertices in our morphable model. More formally, given the generated face $\mathbf{F_g} = \{\mathbf{v_1, v_2,..., v_n}\}$ and face pose $(\mathbf{R, t})$, we find the 2D projection as 

\begin{equation}
\label{eq3}
    \mathbf{L}^{'} = \mathbf{K\cdot(R \cdot F_g+t)} =  \mathbf{\pi} \cdot \mathbf{F_g}
\end{equation}
where $\mathbf{L^{'}}$ is the new set of dense landmarks used for the next iteration of face pose estimation. This converges in just a few iterations. Algorithm \ref{algo1} shows our FPR method in detail.

\section{Results}
\label{results}
In this section, we first discuss the datasets, evaluation metrics, and implementation details for conducting the experiments. We then perform a qualitative and quantitative comparison to several State-Of-The-Art(SOTA) methods to show the effectiveness of our method. Finally, we provide an ablation study on different components to validate the proposed method.

\subsection{Implementation Details}
\label{impl_det}

We perform our experiments on two widely used face datasets, FaceScape \cite{yang2020facescape} and Stirling dataset \cite{Stirling2018}. The FaceScape dataset contains high-resolution scans of hundreds of identities, each with more than 50 high-resolution images captured by DSLR cameras. We downsample them to a lower resolution (512x341) for a fair comparison.  We use 40 randomly selected individuals as our testset. The Stirling dataset contains high-resolution scans of more than 100 people, each with four views. We follow prior work \cite{bai2021riggable} and use 31 individuals as our testset.

\begin{table*}[htbp]
\centering
\begin{tabular}{|l|c|c|c|c|c|c|}
\hline
 & \multicolumn {3}{|c|}{\textbf{FaceScape}} & \multicolumn {3}{|c|}{\textbf{Stirling}} \\
\hline
Methods & \textbf{MSE} & \textbf{Median} & \textbf{M90} & \textbf{MSE} & \textbf{Median} & \textbf{M90} \\
\hline
MVF-Net \cite{wu2019mvf}  & 1.75 & 1.45 & 3.60 & 1.52 & 1.22 & 3.14 \\
DFNRMVS \cite{bai2020deep} & 1.79  & 1.45 & 3.64 & 1.36  & 1.04 & 2.79\\
INORig \cite{bai2021riggable} & 1.54  & 1.29 & 3.09 & 1.20 &0.90 & 2.50 \\
HRN \cite{lei2023hierarchical} & 1.31 & 0.97 & 2.61 & 1.69 & 1.18 & 3.53\\
Ours & \textbf{1.07} & \textbf{0.83} & \textbf{2.27} & \textbf{1.04} & \textbf{0.83} & \textbf{2.07}\\
\hline
\end{tabular}
\caption{Quantitative comparison of face shape estimation with SOTA methods on FaceScape and Stirling datasets. MSE, Median, and max error after rejecting 10\% outliers (M90) are provided over each test set. Our method clearly outperforms the existing  methods on both FaceScape and Stirling datasets. }
\label{tab:table1}
\end{table*}

For error analysis, the predicted meshes are aligned to the ground truth using the Iterative Closest Point (ICP) algorithm. For each point on the ground truth scan, we calculate the point-to-face distance in millimeters by finding the closest triangle in the predicted mesh. From this set of distances, we calculate summary statistics like mean-squared error (MSE), Median and a robust approximation of maximum error which discards 10\% of high error points as outliers (M90). For the face shape prior used during pose estimation, we use 80\% of the training set scans, none of which are a part of our test set. During the 3DMM fitting, we crop the face to include only the frontal region.

\subsection{Comparison to the State-Of-The-Art}

\subsubsection{Qualitative Comparison}
\label{qual_comp}

In this section, we perform a qualitative comparison of the proposed method with four SOTA methods: MVF-Net \cite{wu2019mvf}, DFNRMVS  \cite{bai2020deep}, INORig \cite{bai2021riggable} and HRN \cite{lei2023hierarchical} on FaceScape  \cite{yang2020facescape} and Stirling \cite{Stirling2018} datasets. We show the rendering of example 3D faces generated using different methods along with their error map distribution over the face in Figure \ref{figure:fig4}. HRN \cite{lei2023hierarchical} uses a geometry disentanglement and introduce the hierarchical representation to fulfill detailed face modeling. Although, it captures the high frequency details, the deformation introduces error near head, chin and cheekbones regions as evident from error map in Figure \ref{figure:fig4}. MVF-Net \cite{wu2019mvf} and DFNRMVS  \cite{bai2020deep} trains convolutional neural networks to explicitly enforce multi-view appearance consistency and learns the pose and shape jointly. In contrast, we solve for the pose and shape disjointly, which enables our pipeline to enforce multi-view consistency using accurate multi-view stereo. The qualitative results shown in Figure \ref{figure:fig4} validates our claim. The first and second rows of Figure \ref{figure:fig4} show the output and error map on sample data from FaceScape, while the third and fourth rows show the same on a sample from the Stirling dataset. As evident from the error maps, our output face approximates the ground truth more closely than the other methods.

The integration of dense stereo reconstruction within our method enables the implementation of further non-rigid deformation. This particularly generates shape refinement beyond overly smooth FLAME face space. We upsample the FLAME face and perform the as-rigid-as-possible deformation \cite{amberg2007optimal} followed by nonrigid ICP \cite{sorkine2007rigid}. We perform Taubin smoothing after, since 2-view stereo is still noisy \cite{Taubin_smooth}. We compare our qualitative results with DECA \cite{feng2021learning} and HRN \cite{lei2023hierarchical} as these methods propose to recover higher frequency features beyond 3DMM models. As shown in the figure \ref{figure:facescape_deform} and figure \ref{figure:stirling_deform}, we are able to recover mid-frequency features which looks visually closer to the ground truth.

In contrast to DNN which often downsample input images, stereo techniques excel in extracting features from high-resolution data. Therefore, in Figure \ref{figure:facescape_deform} we show the scenario while we intentionally use two high resolution inputs. Stereo is generally used at a multi-view setting, so we also show where we increase the number of images (50+) to serve as upper-bound. Note that these two are already similar, we thus conclude that a mid-frequency detailed 3D face reconstruction can be achieved with our method.

In order to demonstrate the problem that arises when attempting to solve pose and shape jointly, we perform a qualitative comparison of our proposed method with a SFM/MVS method  \cite{schoenberger2016sfm, schonberger2016pixelwise} in Figure \ref{fig:figure6}. An ambiguity exists between pose and shape, resulting in a stretched-out face when the optimization converges to a low error solution that is incorrect. Our disjoint pose and shape pipeline allows the use of a strong prior on shape while solving pose, and a known pose while solving shape. This results in finding the correct solution that closely matches the ground truth shape.

\subsubsection{Quantitative Comparison}
\label{quant_comp}

In this section, we present a qualitative comparison of the proposed method with SOTA multi-view methods. Table \ref{tab:table1} shows the MSE, Median and M90 error numbers on the FaceScape testset. It can be seen that the proposed method significantly outperforms SOTA methods on all three error metrics, achieving an MSE error of as low as 1.07mm.

Next, We provide the comparison on the Stirling dataset in Table \ref{tab:table1} (right half). Although Stirling images have lower resolution and higher compression than FaceScape images, the proposed method outperforms SOTA methods, demonstrating the effectiveness of the proposed method on different datasets. Also, notice that our method outperforms DFNRMVS \cite{bai2020deep} and INORig  \cite{bai2021riggable}, which use the Stirling data as their training set.

We have also evaluated the error as a cumulative distribution function, as shown in Figure \ref{figure:fig3}, which provides a more complete summary than a single aggregate metric. A horizontal line through the plot at 0.9 is identical to the M90 error metric reported above. The proposed method exhibits a lower error for all fractions of points on the FaceScape dataset (Figure \ref{figure:fig3} left). The Stirling dataset (Figure \ref{figure:fig3} right) plot indicates that even with lower image quality, the proposed method outperforms other methods.

As mentioned in qualitative comparison, our intermediate stereo output can be used to deform the 3DMM output to recover detailed features. In table \ref{tab:table4} we show the error before and after deformation. While the qualitative results appear more visually appealing, we note a modest enhancement in the error for FaceScape dataset.

\begin{table}[htbp]
\centering
\begin{tabular}{|l|c|c|c|}
\hline
\textbf{Stirling} & \textbf{MSE} & \textbf{Median} & \textbf{M90} \\
\hline
Ours (No Deformation)   & 1.04 & 0.83 & 2.07  \\
Ours (Deformed)  & 0.97  & 0.78 & 1.98 \\
\hline
\textbf{FaceScape} & \textbf{MSE} & \textbf{Median} & \textbf{M90} \\
\hline
Ours (No Deformation)   & 1.07 & 0.83 & 2.27  \\
Ours (Deformed)  & 1.06  & 0.82 & 2.26 \\

\hline
\end{tabular}
\caption{Comparison after performing the deformation to 3DMM output using stereo output}
\label{tab:table4}
\end{table}

The proposed method can easily be generalized to 3 and more views. We show the reconstruction error for 3-views in Table \ref{tab:table3}. It can be seen that the proposed method outperforms the existing methods even on 3-views.

\begin{table}[htbp]
\centering
\begin{tabular}{|l|c|c|c|}
\hline
 & \multicolumn {3}{|c|}{\textbf{FaceScape (3-views)}} \\
\hline
Methods & \textbf{MSE} & \textbf{Median} & \textbf{M90} \\
\hline
MVF-Net \cite{wu2019mvf}  & 1.75 & 1.45 & 3.60  \\
DFNRMVS \cite{bai2020deep} & 1.73  & 1.40 & 3.56 \\
INORig \cite{bai2021riggable} & 1.46  & 1.18 & 3.05 \\
HRN \cite{lei2023hierarchical} & 1.28 & 0.96 & 2.54 \\
Ours & \textbf{0.97} & \textbf{0.80} & \textbf{2.01} \\
\hline
\end{tabular}
\caption{Quantitative comparison using 3-views input with SOTA methods on FaceScape datasets.}
\label{tab:table3}
\end{table}

\subsection{Ablation Study}
\label{ablation}
In this section, we provide an ablation study of different components of the proposed pipeline. 
In the pose estimation stage of the pipeline, results are affected by both 2D dense landmark inaccuracies and face prior inaccuracies. Mistakes made estimating pose will propagate to errors in shape reconstruction.  Therefore, we perform an analysis of how the reconstruction accuracy is impacted by changes in these two factors. 


In the first experiment, we aim to find the upper bound of the proposed method if we improve our face shape prior. We replace the mean face prior with an ideal prior, the ground truth 3D scan, when estimating the face pose. It can be observed in Table \ref{table2} that MSE error improves from 1.07 to 1.05mm. However, this is a very small improvement and we conclude that a mean face prior, though simple, is a sufficient prior while estimating pose. 

Next, we perform an analysis to find the upper bound reconstruction accuracy with improvement in 2D landmarks. For this, we use ground truth 2D landmark to find the face pose. In Table \ref{table2}, the first row shows the MSE using the proposed method without FPR, while the last row shows when ground truth landmarks are used, again without FPR. A significant improvement of $\sim$ 0.3mm (from 1.16 to 0.87) is observed when landmark localization is improved. Based on this observation, we introduced the iterative FPR portion of our pipeline to improve landmarks. As shown in Table \ref{table2}, second row, the proposed FPR method does improve accuracy, reducing error to 1.07. However, there is still a scope for improvement.

\begin{figure}

\includegraphics[width=\linewidth]{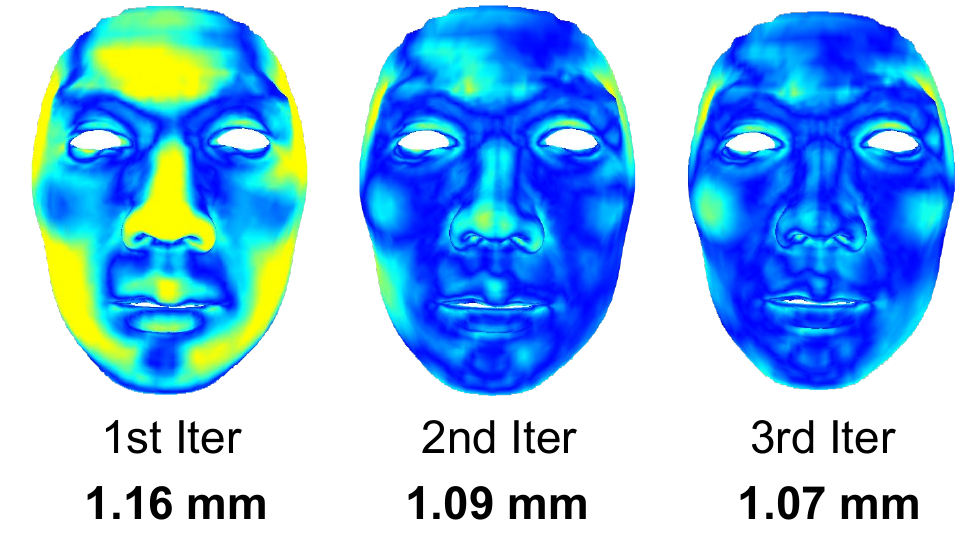}
    \caption{(Top) error map for a sample input and (Bottom) average error on FaceScape dataset after each iteration of FPR. A clear improvement in reconstruction performance is observed using the proposed FPR.}
    \label{fig:figure5}

\end{figure}

If both ground truth 3D face and ground truth 2D landmarks exist it is possible to solve for pose exactly. This is equivalent to pre-calibrated cameras. In this case the remaining error is due to stereo reconstruction and morphable model fitting. However this error is substantially lower at 0.77mm. We conclude that although pose has many fewer parameters than shape, finding correct pose is critical to future progress in 3D face reconstruction.

\begin{table}
\begin{center}
\begin{tabular}{|l|l|l|l|}
\hline
 & Mean Face & GT Face \\
\hline\hline
Ours (No FPR)  & 1.16 & 1.09 \\
Ours (Using FPR after 3 iter) & 1.07 & 1.05\\
GT Landmarks (No FPR) & 0.87 & 0.77 \\
\hline
\end{tabular}
\end{center}
\caption{Ablation study for change in reconstruction performance (MSE in mm) with change in 2D landmarks and face shape prior. It can be observed that using GT 2D landmark provides a significant improvement of \textasciitilde 0.3mm while GT Face provides less than .1 mm improvement. The proposed FPR method reduces error to 1.07mm}
\label{table2}
\end{table}

Finally, we also perform an analysis of how the reconstruction error improves after each iteration of the proposed FPR process. In Figure \ref{fig:figure5}, we show the MSE for the first three iterations on a sample input. It can be observed that the proposed FPR improves the MSE from 1.16 mm to 1.07 mm. The iterative FPR converges in just three iterations due to good initialization and an end-to-end topologically consistent pipeline.

\section{Limitations}
In comparison to the state-of-the-art methods, our proposed method achieves significantly better 3D face reconstruction both quantitatively and qualitatively. Our method's reliance on photometric stereo as an intermediary step means that it faces challenges. For instance, it struggles when there's a significant angle between the input views, leading to difficulties in stereo matching due to large disparity and matching ambiguity. Additionally, the requirement for images to be captured under uniform conditions is essential for effective stereo matching. 

\section{Conclusion}

This paper proposes an end-to-end method for 3D face reconstruction from two uncalibrated images. We introduce a strong face shape prior to the face pose estimation in order to make the optimization stable and accurate. This pose is then used for stereo reconstruction, followed by a 3DMM fitting to find the face shape. An iterative face pose refinement procedure improves the face pose and consequently reconstruction accuracy. The proposed method is evaluated on two widely used face datasets, and outperforms SOTA methods. 
\label{concl}

{\small
\bibliographystyle{ieee_fullname}
\bibliography{main}
}
\end{document}